\documentclass{article}
\usepackage{graphicx}
\usepackage{amsfonts}
\usepackage{amsmath}
\usepackage{amssymb}
\PassOptionsToPackage{hyphens}{url}\usepackage{hyperref}
\usepackage[numbers,super]{natbib}
\usepackage[margin=1in]{geometry}
\usepackage[hang]{footmisc}
\usepackage{url}
\usepackage[colorinlistoftodos]{todonotes}
\usepackage{algorithm}
\usepackage{algpseudocode}
\usepackage{booktabs}
\usepackage{makecell}
\usepackage{pifont}
\newcommand{\cmark}{\ding{51}}
\newcommand{\xmark}{\ding{55}}
\usepackage{upquote}

\input{macros}

\title{GDC Cohort Copilot: An AI Copilot for Curating Cohorts from the Genomic Data Commons}
\author{
Steven Song\textsuperscript{\cofirst,\affilCTDS,\affilCS,\affilMSTP}
\and
Anirudh Subramanyam\textsuperscript{\cofirst,\affilCTDS}
\and
Zhenyu Zhang\textsuperscript{\affilCTDS}
\and
Aarti Venkat\textsuperscript{\affilCTDS,\affilBiomedDS}
\and
Robert L. Grossman\textsuperscript{\corresponding,\affilCTDS,\affilCS,\affilBiomedDS}
}
\date{}

\begin{document}

\maketitle

\renewcommand{\thefootnote}{\myfnsymbol{footnote}}
\footnotetext[1]{Center for Translational Data Science, University of Chicago, Chicago, IL}
\footnotetext[2]{Department of Computer Science, University of Chicago, Chicago, IL}
\footnotetext[3]{Medical Scientist Training Program, Pritzker School of Medicine, University of Chicago, Chicago, IL}
\footnotetext[4]{Section of Biomedical Data Science, Department of Medicine, University of Chicago, Chicago, IL}
\footnotetext[5]{These authors contributed equally}
\footnotetext[6]{Corresponding author: rgrossman1@uchicago.edu}
\setcounter{footnote}{0} 
\renewcommand{\thefootnote}{\alph{footnote}}

\begin{abstract}

\noindent\textbf{Motivation:} The Genomic Data Commons (GDC) provides access to high quality, harmonized cancer genomics data through a unified curation and analysis platform centered around patient cohorts. While GDC users can interactively create complex cohorts through the graphical Cohort Builder, users (especially new ones) may struggle to find specific cohort descriptors across hundreds of possible fields and properties. However, users may be better able to describe their desired cohort in free-text natural language.\\

\noindent\textbf{Results:} We introduce GDC Cohort Copilot, an open-source copilot tool for curating cohorts from the GDC. GDC Cohort Copilot automatically generates the GDC cohort filter corresponding to a user-input natural language description of their desired cohort, before exporting the cohort back to the GDC for further analysis. An interactive user interface allows users to further refine the generated cohort. We develop and evaluate multiple large language models (LLMs) for GDC Cohort Copilot and demonstrate that our locally-served, open-source GDC Cohort LLM achieves better results than GPT-4o prompting in generating GDC cohorts.\\

\noindent\textbf{Availability and implementation:} We implement and share GDC Cohort Copilot as a containerized Gradio app on HuggingFace Spaces, available at \url{https://huggingface.co/spaces/uc-ctds/GDC-Cohort-Copilot}. GDC Cohort LLM weights are available at \url{https://huggingface.co/uc-ctds}. All source code is available at \url{https://github.com/uc-cdis/gdc-cohort-copilot}.

\end{abstract}


\section{Introduction}

The National Cancer Institute's (NCI) Genomic Data Commons (GDC) is a highly used resource for cancer research. With over 100,000 unique monthly users, the GDC provides access to high quality, harmonized, multimodal cancer data for over 45,000 patient cases\citep{heath2021nci}. A typical user workflow using the GDC is to curate a cohort of cases before doing subsequent analysis, either using tools within the GDC Data Portal or through the GDC API\citep{jensen2017nci}. Central to this workflow is the set of filters used to construct the cohort.

The GDC provides the Cohort Builder tool to allow users to interactively select their desired filters. The Cohort Builder is a powerful tool which allows users to select specific values from over 700 filter properties. While the Cohort Builder organizes commonly used filters into user-readable groupings, there are still dozens of properties, each with potentially a hundred or more possible values to filter by. This balance of allowing users to create specific and verbose filters while providing a user-friendly interface is complex. New users of the GDC may especially find it difficult to identify the filters most relevant for their research. However, such users may naturally be able to describe their desired cohort in natural language.

Here, we present GDC Cohort Copilot, an open-source AI copilot that enables users to curate GDC cohorts using natural language. Following the recent success of large language models (LLMs) in generating structured code\cite{chen2021evaluating} and database query languages\cite{pourreza2025chase, ganesan2024llm}, the GDC Cohort Copilot is powered by GDC Cohort LLM, an LLM trained to generate structured GDC cohort filters from free-text user input. We demonstrate that our locally-served, open-source model outperforms GPT-4o prompting in cohort construction accuracy. Once generated by the model, the tool automatically populates the cohort filter into a GDC Cohort Builder-like interface that allows the user to further refine their desired cohort. We provide a mechanism for exporting the curated cohort back to the GDC for further analysis. We release GDC Cohort Copilot as the overall framework presented in Figure \ref{fig:overview}, the GDC Cohort LLM, and the containerized web app.

\section{Materials and methods}
\begin{figure}[t]
    \centering
    \includegraphics[width=6in,height=4in]{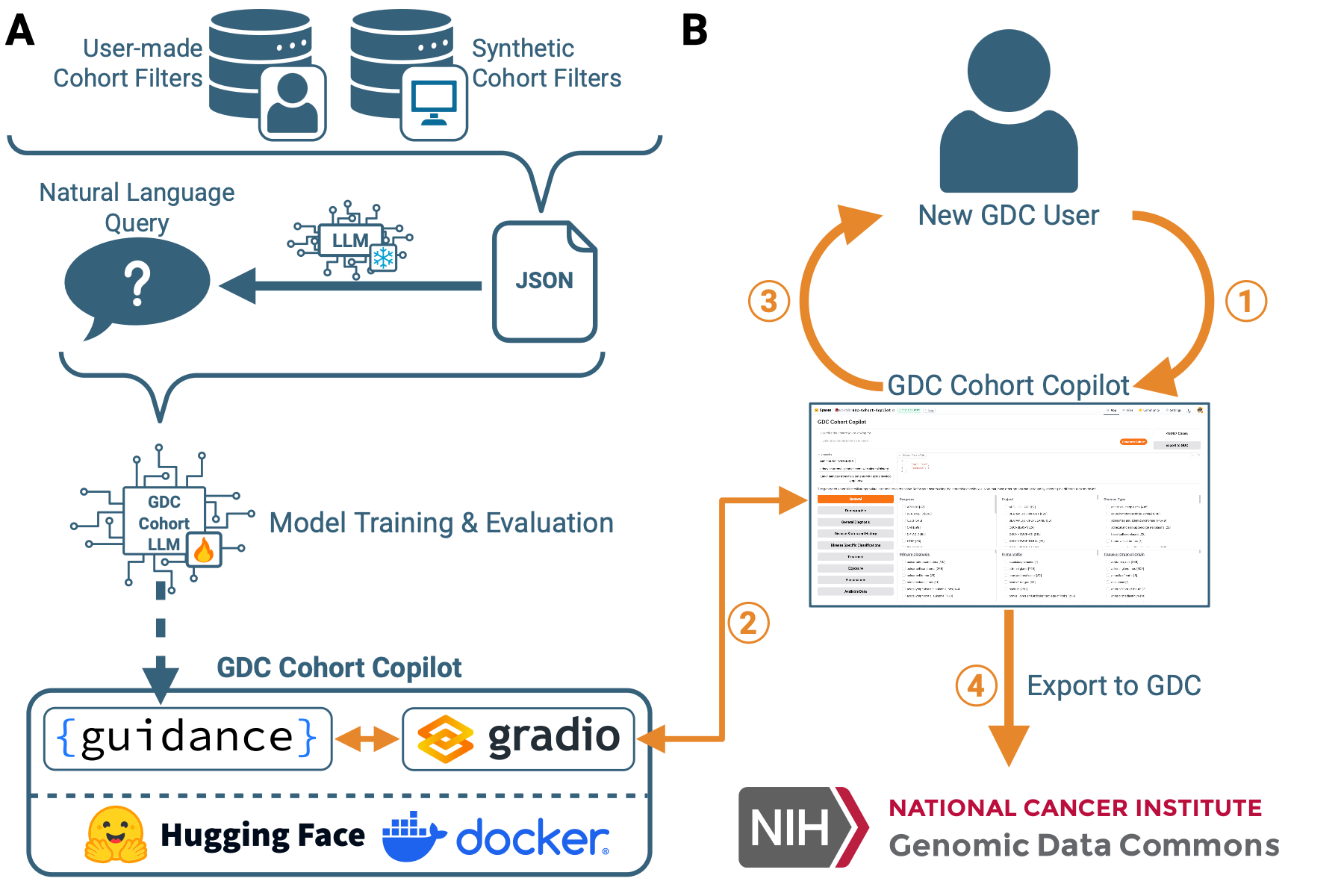}
    \caption{Overview of GDC Cohort Copilot implementation and user workflow. (A) Implementation of GDC Cohort Copilot involves training the GDC Cohort LLM to translate from a natural language query of a cohort to the cohort filter JSON. The cohort JSONs are derived from datasests of real user-made cohorts or synthetically generated cohorts. The paired natural language queries are generated by a frozen LLM using the cohort JSONs. The final trained GDC Cohort LLM model is served in a containerized web app that exposes a GDC Cohort Builder-like interface running on HuggingFace Spaces. (B) A user curates their desired cohort using the GDC Cohort Copilot by: (1) inputting a natural language description of a desired cohort (2) which is automatically passed to GDC Cohort LLM. The model is served using Guidance inside a Gradio app. (3) The resulting generated cohort filter is automatically populated back into the interface, allowing the user to manually refine their cohort before (4) exporting the curated cohort to the NCI GDC.}
    \label{fig:overview}
\end{figure}

GDC Cohort Copilot is comprised of both the generative GDC Cohort LLM and the containerized web app interface. The overview of its implementation and user workflow is presented in Figure \ref{fig:overview}.

\begin{table}[t]
\centering
\begin{tabular}{lrrrr}
\toprule
Model      & TPR   & IoU   & Exact & BERT \\
\midrule
\textbf{GDC Cohort LLM} & \textbf{0.855} & \textbf{0.832} & \textbf{0.702} & \textbf{0.919} \\
GPT-4o     & 0.720 & 0.698 & 0.558 & 0.894 \\
\bottomrule
\end{tabular}

\caption{GDC Cohort LLM is significantly better at generating GDC cohorts across all reported metrics compared to GPT-4o ($p < 0.05$). TPR: true positive rate; IoU: intersection over union; Exact: exact match; BERT: F1 BERTScore using SciBERT.}
\label{tab:comparison}
\end{table}

\subsection{Inputs and Outputs}
The primary input to the GDC Cohort Copilot is a natural language description of a GDC cohort, for example: ``cases with gene expression data derived from RNA sequencing for lung adenocarcinoma''. Upon submitting the query, the app uses GDC Cohort LLM model to generate and return the corresponding cohort filter JSON. The interface automatically populates the corresponding checkboxes for filter properties specified by the generated JSON. A user can interactively refine the cohort selections, before ultimately exporting and outputting a text file of GDC case identifiers. These case identifiers can be imported by the GDC for further analysis.

\subsection{Core Set of Filter Properties}
\label{sec:core-set}
In this initial release of the GDC Cohort Copilot, we simplify the development of the tool by considering only a subset of 68 filter properties from the GDC Cohort Builder. These are the default and most commonly used filter properties exposed by the GDC Data Portal v2.4.0 and additionally have predefined lists of possible values or value ranges (e.g. disease type or age at diagnosis). We refer to this subset of filters as the ``core set''. We create a JSON schema (from a Pydantic data model) to validate possible filters comprised of the core set.

\subsection{GDC Cohort LLM}
The GDC Cohort Copilot is powered by a generative LLM, GDC Cohort LLM, which translates natural language queries of cohorts into cohort filter JSONs. We describe here the development and evaluation of GDC Cohort LLM.

\subsubsection{Data}
\label{sec:data}
GDC Cohort LLM is trained over paired natural language queries and cohort filter JSONs. This data is derived from real user-generated and synthetic cohort filters. 68,209 user-generated cohort filters were supplied by the GDC User Services team from their database of GDC user-saved cohorts. Removing duplicates, null filters, filters with properties outside of the core set, and filters which fail schema validation results in 16,235 usable cohort filters. We additionally experiment with augmenting our dataset by randomly sampling synthetic cohorts filters. Specifically, we randomly sample fields and values from the core set of filter properties. Further details on our random sampling procedure are provided in Section \ref{sec:random-cohort}. We experiment with augmenting our training data using 100,000 and 1,000,000 synthetic cohort filters.

One of the primary limitations of our cohort filter dataset is that it does not contain any user-generated natural language descriptions of the cohorts. To address this challenge, we prompt Mistral-7B-Instruct-v0.3\citep{jiang2023mistral7b} to generate a corresponding natural language query for a given cohort filter JSON. Our precise procedure for this reverse translation, including the prompt we use, is provided in Section \ref{sec:reverse-translation}. We apply this method to all real and synthetic cohort filters.

We finally split our paired samples derived from real user-generated data into 14,235 for training and 2,000 for evaluation. For the evaluation split, we ensure that the model-specific token length of the natural language query and cohort filter for all samples fit within the minimum context length of the different models we experiment with. Additionally, we require that the evaluation samples do not result in empty cohorts (cohorts with 0 cases). This allows us to directly compare each experiment's results which were derived over precisely the same set of data samples.

\subsubsection{Model}
\label{sec:model}
We experiment with three pretrained LLMs of different architectures and scales: GPT-2\citep{radford2019language}, BART\citep{lewis-etal-2020-bart}, and Mistral-7B-Instruct-v0.3\citep{jiang2023mistral7b}. We train each of these models using a causal language modeling (CLM - autoregressive) objective. For BART, the input to the encoder is the natural language query while the output of the decoder is the cohort filter JSON. For GPT-2 and Mistral, we concatenate the natural language query with the cohort filter JSON. Additionally, for Mistral, we use low rank adaptation (LoRA)\citep{hu2022lora} to efficiently train the model for our translation task. We load model weights from HuggingFace and use HuggingFace utilities for training our models. Further training details are described in Section \ref{sec:training}.

At evaluation time, for efficient batched inferencing, we serve the trained models using vLLM\citep{kwon2023efficient} with structured decoding using Outlines\citep{willard2023efficient} to ensure that our generated outputs are valid cohort filter JSON. One limitation of the JSON schema we develop (Section \ref{sec:core-set}) is that it does not strictly enforce field and property strings; rather, our schema enforces the structure of the filter.

\subsubsection{Evaluation}
\label{sec:evaluation}
After training the variations of GDC Cohort LLM, differing either in model type or data mixture, we evaluate the generated cohort filters. As we aim to enable accurate retrieval of cohorts, we do not directly evaluate the cohort filter; instead we compare the cases retrieved by the generated cohort filter to the cases retrieved by the true cohort filter. This allows flexibility in the actual content of the cohort filter as many filters may result in the same set of cases, for example selecting the TCGA program is equivalent to selecting all of the individual TCGA projects together. We thus compute three metrics for each filter's set of cases: sensitivity (true positive rate - TPR), Jaccard index (intersection over union - IoU), and a binary indicator for if the predicted and actual cases precisely match (Exact). If a generated filter is not valid (either due to context length truncation or imprecise generated field or value names) and cannot be used to retrieve cases using the GDC API, we use the empty set. TPR, IoU, and Exact are guaranteed to be finite and equal 0 if there are no predicted cases, as we ensure that the actual cases are never null (Section \ref{sec:data}).

While we do not require the generated cohort filter precisely match the actual filter, we do evaluate whether they are semantically similar. To do so, we reverse translate the predicted cohort filters into natural language queries and compute the F1 BERTScore\citep{zhang2020bert} (BERT) between the original and derived queries. We specifically use SciBERT\citep{beltagy2019scibert} in the computation of the BERTScore. Additionally, we compare all of our trained models against a prompting-based alternative using OpenAI's GPT-4o\cite{hurst2024gpt} that requires an expensive, long context window of approximately 15,000 tokens per prompt. Further details of this comparison implementation are provided in Section \ref{sec:gpt4-prompting}. Finally, we report the average of all metrics across all 2,000 evaluation filters and apply paired t-tests (for TPR, IoU, BERT metrics) or McNemar's test (for Exact metric) with Bonferroni correction to evaluate statistical significance.

\subsection{Web App, Containerization, and Deployment}

We develop the web app for GDC Cohort Copilot as a Gradio\cite{abid2019gradio} app deployed in a HuggingFace Space. HuggingFace Spaces provides out-of-the-box containerization with Gradio apps, enabling users to download and run GDC Cohort Copilot locally with docker. We package the GPT-2 variant of GDC Cohort LLM, trained over real and 1 million synthetic data samples, with GDC Cohort Copilot; in addition to its strong evaluation metrics, its architecture as a decoder-only, small-scale LLM enables it to be efficiently served. Specifically, we serve GDC Cohort LLM using Guidance\footnote{https://github.com/guidance-ai/guidance} for structured generation. While we utilize GPU acceleration in our HuggingFace Space for serving GDC Cohort LLM, the model only requires approximately 1 GB of GPU VRAM and can even run efficiently on CPU. Our implementation of GDC Cohort Copilot allows it to be accessible to a wide variety of biomedical research users.

\begin{table}[t]
\centering
\begin{tabular}{lcc}
\toprule
Comparison & GDC Cohort LLM & GPT-4o \\
\midrule
Achieves SOTA      & \cmark & \xmark \\
Open source        & \cmark & \xmark \\
Deploy locally     & \cmark & \xmark \\
Structured outputs & \cmark & \cmark \\
No training        & \xmark & \cmark \\
Required tokens    & $\leq 1024$ & $>15k$   \\
\bottomrule
\end{tabular}
\caption{Conceptual comparison of GDC Cohort LLM and GPT-4o as LLMs to power GDC Cohort Copilot.}
\label{tab:conceptual}
\end{table}

\section{Results}

We first evaluate the adaptability of various model types to our filter generation task as GDC Cohort LLM (Table \ref{tab:model_arch}). Training over user-derived data, we find that GPT-2 (TPR=0.365; IoU=0.331; Exact=0.221) significantly outperforms BART (TPR=0.117, p=8.56e-89; IoU=0.078, p=7.73e-114; Exact=0.028, p=3.69e-94) and Mistral (TPR=0.124, p=2.33e-90; IoU=0.117, p=3.62e-80; Exact=0.092, p=9.35e-39) models over case-retrieval metrics. Over query-based metrics, GPT-2 (BERT=0.819) outperforms BART (BERT=0.735, p=4.03e-106). While GPT-2 is statistically worse than Mistral (BERT=0.835, p=6.47e-5), the difference is relatively small and not meaningful in the context of poor case-retrieval capabilities.

Given GPT-2's strong adaptability to GDC Cohort LLM, we next explore how to improve its performance by training over synthetic data mixtures (Table \ref{tab:data_mix}). We find that, compared to a baseline using only user-derived data (TPR=0.365; IoU=0.331; Exact=0.221; BERT=0.819), incorporating 100 thousand synthetically generated records with our real user data (TPR=0.783, p=9.59e-217; IoU=0.748, p=7.19e-227; Exact=0.607, p=1.00e-188; BERT=0.902, p=7.67e-145) significantly improves all metrics. We further train over a mixture of 1 million synthetic records with user records and find that this provides significantly stronger results (TPR=0.855, p=6.03e-18; IoU=0.832, p=1.85e-23; Exact=0.702, p=1.20e-23; BERT=0.919, p=3.74e-16) than using only 100 thousand synthetic samples.

Our final GDC Cohort LLM model is thus trained from a GPT-2 foundation over a mixture of 1 million synthetic and real user data. Importantly, GDC Cohort LLM (TPR=0.855; IoU=0.832; Exact=0.702; BERT=0.919) significantly outperforms a prompting-based implementation of cohort filter generation using GPT-4o (TPR=0.720, p=8.01e-37; IoU=0.748, p=2.08e-36; Exact=0.607, p=2.12e-37; BERT=0.894, p=3.57e-26) across all metrics (Table \ref{tab:comparison}). Because GDC Cohort LLM is specifically trained for this task, to provide a more fair comparison, we prompt GPT-4o with a list of all possible field-value pairs which consume 15K tokens. This reduces the potential for GPT-4o to hallucinate invalid field or value names. Despite this, we find that our open-source, small-scale GDC Cohort LLM model achieves better results than GPT-4o. We conceptually compare GDC Cohort LLM to GPT-4o in Table \ref{tab:conceptual}.

Finally, we package GDC Cohort LLM with our GDC Cohort Copilot tool as a containerized Gradio app running on HuggingFace Spaces. GDC Cohort Copilot can additionally be downloaded and run locally using docker. We serve a GDC Cohort Builder-like interface to allow users to interactively curate cohorts using both natural language based descriptions and graphical checkboxes. We integrate with NCI GDC by providing utilities to export curated cohorts back to the GDC for further user-specific analysis.

\section{Conclusion}

GDC Cohort Copilot is a novel copilot tool to use natural language to assist in the curation of cohorts from the NCI GDC. Users can interactively use the copilot and a graphical interface to discover and refine their cohort. We experiment with various model types and data mixtures to develop GDC Cohort LLM, and demonstrate that our open-source, small-scale model is better able to accurately translate natural language descriptions of cohorts into their corresponding GDC cohort filter. We share GDC Cohort Copilot as a containerized Gradio app deployed on HuggingFace Spaces, ultimately providing an accessible tool to aid biomedical cancer researchers in their data curation efforts.

\clearpage
\section*{Acknowledgements}

This work was funded in part through the Advanced Research Projects Agency for Health (ARPA-H) under contract 75N92020D00021/5N92023F00002. The views and conclusions contained in this document are those of the authors and should not be interpreted as representing the oﬃcial policies, either expressed or implied, of the U.S. Government. S.S. is additionally supported by NIH training grant T32GM007281. Overview figure was created using BioRender.

\makeatletter
\renewcommand \thesection{S\@arabic\c@section}
\renewcommand\thetable{S\@arabic\c@table}
\renewcommand \thefigure{S\@arabic\c@figure}
\makeatother
\setcounter{section}{0}
\section{Supplemental information}


\subsection{Supplemental Methods}

\subsubsection{Random Cohort Generation}
\label{sec:random-cohort}
We augment our existing user-derived cohort data by creating synthetic cohort filter JSONs through a two-step stochastic process. To create a new JSON filter, we first select the number of fields, $n$, to include by sampling a nonzero integer value from a Chi-Square distribution with 6 degrees of freedom, using rejection-sampling to ensure i.i.d. samples. Next, we randomly select $n$ distinct fields from the core set of filter properties. For fields with numerical ranges, we randomly select a filter operator (one of $\leq$, $<$, $\geq$, $>$) and a corresponding numerical value within the field's allowable range. Conversely, all non-numerical fields use an ``in'' operator with a list of values. To create this list of filtering values for a given field, we first randomly select the number of values, $m$, with a minimum of 1 and a maximum of 5. The $m$ values are then randomly selected from the list of possible values for the given field. We construct each filter as a JSON object and validate it against the GDC cohort schema to ensure correctness. Synthetic filter uniqueness is strictly enforced by computing an MD5 hash digest of the sorted JSON representation and rejecting all duplicates. We iteratively generate filters until the target number of unique, valid, synthetic cohort filters are generated.

\subsubsection{Reverse Translation}
\label{sec:reverse-translation}
In our dataset of real user-derived cohort filters (originally, assembled using the GDC Cohort Builder), the cohort filters are JSON strings. As we aim to train an LLM to generate these filters using natural language, we develop a reverse-translation method to prompt an LLM to generate a JSON filter's corresponding natural language description. Specifically, we use \texttt{mistralai/Mistral-7B-Instruct-v0.3} for this task without any additional fine-tuning. We serve the model using vLLM on a single NVIDIA A100 80GB GPU and generate natural language queries using a set seed, temperature 0, and a maximum of 4,096 completion tokens. We use in-context learning to prompt the model with 2 examples of our task; our precise prompt for reverse-translation is provided in Table \ref{tab:reverse_prompt}. We use this same process to generate natural language queries for our synthetic data samples (Section \ref{sec:random-cohort}). Additionally, we use this method in our evaluation of semantic similarity between two cohort filters (Section \ref{sec:evaluation}).

\subsubsection{Model Training Details}
\label{sec:training}
We describe here our model training details for our experiments in developing GDC Cohort LLM. All models were trained using a single NVIDIA A100 80GB GPU. For all experiments, we use a linear learning rate schedule with an initial learning rate of 5e-5, 8 gradient accumulation steps, and a maximum sequence length of 1,024. For training the LoRA adapter for Mistral, we target the following layers: \texttt{q\_proj, k\_proj, v\_proj, o\_proj, gate\_proj, down\_proj, up\_proj}. Specific hyperparameters for each experiment are further detailed in Table \ref{tab:training}.

\subsubsection{Comparison to GPT-4o}
\label{sec:gpt4-prompting}
To compare our trained GDC Cohort LLM against an expensive, closed source alternative using OpenAI's GPT-4o, we adopt the following prompting strategy. We use the OpenAI python library v1.84.0 to send requests to the OpenAI API. We use OpenAI's Structured Outputs by specifying our Pydantic data model (Section \ref{sec:core-set}) as the response format to chat completions. This same data model and corresponding JSON schema are also used by our locally served model. We pin the model version to \texttt{gpt-4o-2024-08-06}, set temperature to 0, set a fixed generation seed, and specify a maximum of 1,024 completion tokens. We use the prompt presented in Table \ref{tab:openai_prompt}. To provide GPT-4o with the set of all possible filter fields and their values without model tuning, we input in the prompt the entire field-value mapping for the core set of filter properties. This mapping is approximately 15,000 tokens long and thus, regardless of performance metrics, this method is infeasible for scalable inference as the token cost rapidly compounds. At the time of our experiments, the cost of our prompt to GPT-4o was approximately \$0.04.

\begin{table}[t]
\centering
\begin{tabular}{|p{5.75in}|}
\toprule
{\small\begin{verbatim}
Given the following examples of dict and sentence pairs, generate the 
sentence that describes a new dict between <<>>.
Use the 'field' and it's corresponding 'value' information to correctly 
identify the different categories.
Examples:

Example 1:
Dict :
{'op': 'and',
 'content': [{'op': 'in',
   'content': {'field': 'cases.project.program.name', 'value': ['TARGET']}},
  {'op': 'in',
   'content': {'field': 'cases.project.project_id',
    'value': ['TARGET-ALL-P1',
     'TARGET-ALL-P2',
     'TARGET-ALL-P3',
     'TARGET-AML']}},
  {'op': 'in',
   'content': {'field': 'cases.diagnoses.site_of_resection_or_biopsy',
    'value': ['bone marrow']}},
  {'op': 'in',
   'content': {'field': 'cases.samples.tissue_type', 'value': ['tumor']}},
  {'op': 'in',
   'content': {'field': 'cases.samples.tumor_code',
    'value': ['acute lymphoblastic leukemia (all)']}}]}
Sentence : 
acute lymphoblastic leukemia tumor code for bone marrow tumors that belong to 
TARGET-ALL-P1, TARGET-ALL-P2, TARGET-ALL-P3, TARGET-AML projects. |<eos>|

Example 2:
Dict:
{'op': 'and',
 'content': [{'op': 'in',
   'content': {'field': 'cases.project.program.name', 'value': ['CGCI']}},
  {'op': 'in',
   'content': {'field': 'cases.project.project_id', 'value': ['CGCI-BLGSP']}},
  {'op': 'in',
   'content': {'field': 'cases.diagnoses.tissue_or_organ_of_origin',
    'value': ['hematopoietic system, nos']}},
  {'op': 'in',
   'content': {'field': 'cases.samples.preservation_method',
    'value': ['ffpe']}}]}
Sentence:
ffpe samples for hematopoietic system, nos that belong to the CGCI-BLGSP 
project. |<eos>|

\end{verbatim}
\verb|<<|\textcolor{red}{\{Filter\}}\verb|>>|
\begin{verbatim}
    
Sentence: 
\end{verbatim}}\\
\bottomrule
\end{tabular}

\caption{Prompt to reverse-translate natural language descriptions of input cohort filters using an LLM.}
\label{tab:reverse_prompt}
\end{table}

\begin{table}[t]
\centering
\renewcommand{\arraystretch}{1.5}
\begin{tabular}{lrrrrr}
\toprule
Model      & Trainable Params & Train Epochs   & Batch Size   & Precision & Specifics\\
\midrule
Mistral     &  11M & 1 & 64 & bf16 & \makecell[r]{LoRA(r=4, $\alpha$=16,\\ dropout=0.1)} \\
BART        & 139M & 3 & 32 & bf16 &  warmup steps=0 \\
GPT-2       & 124M & 10 & 32 & fp16 & warmup steps=500 \\
GPT-2-100K  & 124M & 10 & 32 & fp16 & warmup steps=500 \\
GPT-2-1M    & 124M & 10 & 32 & fp16 & warmup steps=500 \\
\bottomrule
\end{tabular}

\caption{Model training hyperparameters for GDC Cohort LLM experiments.}
\label{tab:training}
\end{table}

\begin{table}[t]
\centering
\begin{tabular}{llp{5in}}
\toprule
Line & Role & Message \\
\midrule
1 & System & Construct NCI GDC cohort filters based on the input cohort description using the given list of possible fields and values. \\
\midrule
2 & User & \makecell[l]{Here is the list of possible fields and values: \\\\{}[Field-Value List]\\\\Use the above properties to construct a NCI GDC cohort filter for the following \\ cohort description:\\{}[Query] } \\
\bottomrule
\end{tabular}

\caption{Prompt to derive cohort filters using OpenAI's GPT-4o. The field-value list is populated using a predefined mapping for all core set filter properties. This list consumes approximately 15,000 tokens. Cohort filters are generated using structured decoding.}
\label{tab:openai_prompt}
\end{table}


\clearpage
\subsection{Supplemental Results}

\begin{table}[h]
\centering
\begin{tabular}{lrrrr}
\toprule
Model Type & TPR   & IoU   & Exact & BERT \\
\midrule
GPT-2      & \textbf{0.365} & \textbf{0.331} & \textbf{0.221} & 0.819 \\
BART       & 0.117 & 0.078 & 0.028 & 0.735 \\
Mistral    & 0.124 & 0.117 & 0.092 & \textbf{0.835} \\
\bottomrule
\end{tabular}

\caption{GPT-2 significantly outperforms ($p < 0.05$), or is not meaningfully worse than, BART and Mistral as foundations for GDC Cohort LLM, when trained over user-derived cohort filters.}
\label{tab:model_arch}
\end{table}

\begin{table}[h]
\centering
\begin{tabular}{lrrrr}
\toprule
Data Mixture          & TPR   & IoU   & Exact & BERT \\
\midrule
Real                  & 0.365 & 0.331 & 0.221 & 0.819 \\
Real + 100K Synthetic & 0.783 & 0.748 & 0.607 & 0.902 \\
Real + 1M Synthetic   & \textbf{0.855} & \textbf{0.832} & \textbf{0.702} & \textbf{0.919} \\
\bottomrule
\end{tabular}

\caption{Training over synthetic data mixtures significantly improves GDC Cohort LLM performance ($p < 0.05$).}
\label{tab:data_mix}
\end{table}

\clearpage
\bibliographystyle{unsrtnat}
\bibliography{references}

\end{document}